\def\year{2022}\relax
\documentclass[letterpaper]{article} % DO NOT CHANGE THIS
\usepackage{aaai22}  % DO NOT CHANGE THIS
\usepackage{times}  % DO NOT CHANGE THIS
\usepackage{helvet}  % DO NOT CHANGE THIS
\usepackage{courier}  % DO NOT CHANGE THIS
\usepackage[hyphens]{url}  % DO NOT CHANGE THIS
\usepackage{graphicx} % DO NOT CHANGE THIS
\usepackage{natbib}  % DO NOT CHANGE THIS AND DO NOT ADD ANY OPTIONS TO IT
\usepackage{caption} % DO NOT CHANGE THIS AND DO NOT ADD ANY OPTIONS TO IT
\DeclareCaptionStyle{ruled}{labelfont=normalfont,labelsep=colon,strut=off} % DO NOT CHANGE THIS
\usepackage{algorithm}
\usepackage{placeins}
\usepackage{algorithmicx}
\usepackage{tabularx}
\usepackage{graphicx}
\usepackage{placeins}
\usepackage{centernot}
\usepackage{siunitx}
\usepackage{xspace}
\usepackage{amsmath}
\usepackage{hyperref}
\usepackage{booktabs}
\usepackage{multirow}
\usepackage{subcaption}
\usepackage{makecell}
\usepackage[noend]{algpseudocode}

\usepackage{newfloat}
\usepackage{listings}
\floatstyle{ruled}
\newfloat{listing}{tb}{lst}{}
\floatname{listing}{Listing}
\title{GePA*SE: Generalized Edge-Based Parallel A* for Slow Evaluations}

\author{
Shohin Mukherjee, Maxim Likhachev
}
\affiliations{
    The Robotics Institute, CMU\\
    \{shohinm, mlikhach\}@andrew.cmu.edu\\
}

\usepackage{bibentry}
\newcommand{\astar}{A*\xspace}
\newcommand{\wastar}{w\astar\xspace}
\newcommand{\pase}{PA*SE\xspace}
\newcommand{\wpase}{\mbox{w}\pase}
\newcommand{\epase}{ePA*SE\xspace}
\newcommand{\gepase}{GePA*SE\xspace}
\newcommand{\wgepase}{\mbox{w-}\gepase}

\newcommand{\wepase}{\mbox{w-}\epase}

\newcommand{\state}{\ensuremath{\mathbf{s}}\xspace}

\newcommand{\States}{\ensuremath{\mathcal{S}}\xspace}
\newcommand{\startstate}{\ensuremath{\state_0}\xspace}
\newcommand{\ac}{\ensuremath{\mathbf{a}}\xspace}
\newcommand{\dac}{\ensuremath{\mathbf{a^d}}\xspace}
\newcommand{\pac}{\ensuremath{\mathbf{a^d}}\xspace}
\newcommand{\Aset}{\ensuremath{\mathcal{A}}\xspace}
\newcommand{\Asetcheap}{\ensuremath{\Aset^{c}}\xspace}
\newcommand{\Asetexp}{\ensuremath{\Aset^{e}}\xspace}
\newcommand{\goalreg}{\ensuremath{\mathcal{G}}\xspace}
\newcommand{\ed}{\ensuremath{e}\xspace}

\newcommand{\edge}{\ensuremath{(\mathbf{s},\mathbf{a})}\xspace}
\newcommand{\edgeprime}{\ensuremath{(\mathbf{s}',\mathbf{a})}\xspace}
\newcommand{\pedge}{\ensuremath{(\mathbf{s},\mathbf{a^d})}\xspace}
\newcommand{\pedgenext}{\ensuremath{(\mathbf{s}',\mathbf{a^d})}\xspace}
\newcommand{\open}{\ensuremath{\textit{OPEN}}\xspace}
\newcommand{\closed}{\ensuremath{\textit{CLOSED}}\xspace}
\newcommand{\be}{\ensuremath{\textit{BE}}\xspace}
\newcommand{\gval}{\ensuremath{g}}

\newcommand{\cost}{\ensuremath{c}\xspace}
\newcommand{\costopt}{\ensuremath{\cost^*}\xspace}
\newcommand{\fval}{\ensuremath{f}}
\newcommand{\hval}{\ensuremath{h}}
\newcommand{\wh}{\ensuremath{w}\xspace}
\newcommand{\wi}{\ensuremath{\epsilon}\xspace}
\newcommand{\numexpanded}{\ensuremath{n\_successors\_generated}\xspace}
\newcommand{\graph}{\ensuremath{G}\xspace}
\newcommand{\vertex}{\ensuremath{v}\xspace}
\newcommand{\Vertices}{\ensuremath{\mathcal{V}}\xspace}
\newcommand{\Edges}{\ensuremath{\mathcal{E}}\xspace}
\newcommand{\Edgescheap}{\ensuremath{\Edges^{c}}}
\newcommand{\Edgesexp}{\ensuremath{\Edges^{e}}}
\newcommand{\plan}{\ensuremath{\pi}\xspace}
\newcommand{\numthreads}{\ensuremath{N_t}\xspace}

\newcommand{\compratio}{r^c\xspace}

\newcommand{\place}{\textsc{Place}\xspace}

\newcommand{\makeblue}[1]{\underline{#1}}
\newcommand{\domaingrid}{2D Grid World\xspace}
\newcommand{\domainmanip}{Manipulation\xspace}
\begin{document}

\maketitle

\begin{abstract}
Parallel search algorithms have been shown to improve planning speed by harnessing the multithreading capability of modern processors. One such algorithm \pase achieves this by parallelizing state expansions, whereas another algorithm \epase achieves this by effectively parallelizing edge evaluations. \epase targets domains in which the action space comprises actions with expensive but similar evaluation times. However, in a number of robotics domains, the action space is heterogenous in the computational effort required to evaluate the cost of an action and its outcome. Motivated by this, we introduce \gepase: Generalized Edge-based Parallel A* for Slow Evaluations, which generalizes the key ideas of \pase and \epase i.e. parallelization of state expansions and edge evaluations respectively. This extends its applicability to domains that have actions requiring varying computational effort to evaluate them. 
The open-source code for \gepase along with the baselines is available here:\break [Link anonymized] 

% \\\url{https://github.com/shohinm/parallel_search}
\end{abstract}

% \FloatBarrier
\section{Introduction}
Graph search algorithms such as A* and its variants are widely used in robotics for planning problems which can be formulated as a shortest path problem on a graph~\cite{kusnur2021planning, mukherjee2021reactive}. Parallelized graph search algorithms have shown to be effective in robotics domains where action evaluation tends to be expensive. \pase~\cite{phillips2014pa} is an optimal parallel graph search algorithm that strategically parallelizes state expansions to speed up planning, without the need for state re-expansions. However, just like \astar, \pase evaluates all edges at once for every state expansion in the same thread. In contrast, a parallelized planning algorithm \epase (Edge-based Parallel A* for Slow Evaluations)~\cite{mukherjee2022epase} changes the basic unit of search from state expansions to edge expansions. This decouples the evaluation of edges from the expansion of their common parent state, giving the search the flexibility to figure out what edges need to be evaluated to solve the planning problem. In domains with expensive edges, \epase achieves lower planning times and evaluates fewer edges than \pase. \epase is efficient in domains where the action space is homogenous in computational effort i.e. all actions have similar evaluation times. In some domains, the action space can comprise a combination of cheap and expensive to evaluate actions. For example, in planning on manipulation lattices, the action space comprises static primitives generated offline, each of which moves a single joint and Adaptive Motion Primitives, which use an optimization-based IK solver to compute a valid goal configuration (based on the workspace goal) and then linearly interpolate from the expanded state to the goal ~\cite{cohen2014single}. The latter are generated online and are significantly more expensive to compute than the static primitives. In such domains, it is not efficient to delegate a new thread for every edge.
% and parallelization of state expansions similar to \pase is more appropriate.
 
Motivated by these insights, we introduce \gepase which generalizes the key ideas in \pase and \epase i.e. state expansions and edge evaluations respectively. We show that \gepase outperforms both \pase and \epase in domains with heterogenous action spaces by employing a parallelization strategy that handles cheap edges similar to \pase and expensive edges similar to \epase. Additionally, it uses a more efficient strategy to carry out edge independence checks for large graphs. While \gepase is optimal, its bounded suboptimal variant \wgepase inherits the bounded suboptimality guarantees of \wpase and \wepase and achieves faster planning by employing an inflation factor on the heuristic. We evaluate \wgepase in a 2D grid-world and a 7-DoF manipulation domain and demonstrate that it achieves lower planning times in both.

% \FloatBarrier
\section{Related Work}
Several approaches parallelize sampling-based planning algorithms in which multiple processes cooperatively build a PRM~\cite{jacobs2012scalable} or an RRT~\cite{devaurs2011parallelizing, ichnowski2012parallel, jacobs2013scalable}  by sampling states in parallel. However, in many planning domains, sampling of states is not trivial. One such class of planning domains is simulator-in-the-loop planning, which uses a physics simulator to generate successors~\cite{liang2021search}. Unless the state space is simple such that the sampling distribution can be scripted, there is no principled way to sample meaningful states that can be realized in simulation.

We focus on search-based planning which constructs the graph by recursively applying a set of actions from every state. A* can be trivially parallelized by generating successors in parallel when expanding a state. However, the performance improvement is bounded by the branching factor of the domain. Another approach is to expand states in parallel while allowing re-expansions to account for the fact that states may get expanded before they have the minimal cost~\cite{irani1986parallel, evett1995massively, zhou2015massively}. However, they may encounter an exponential number of re-expansions, especially if they employ a weighted heuristic. \pase~\cite{phillips2014pa} parallelly expands states at most once, in such a way that does not affect the bounds on the solution quality. \epase~\cite{mukherjee2022epase} improves \pase by changing the basic unit of the search from state expansions to edge expansions and then parallelizing this search over edges. MPLP~\cite{mukherjee2022mplp} achieves faster planning by running the search lazily and evaluating edges asynchronously in parallel, but relies on the assumption that a successor can be generated without evaluating the edge. There has also been some work on parallelizing A* on GPUs~\cite{zhou2015massively, he2021efficient}. These algorithms have a fundamental limitation that stems from the SIMD (single-instruction-multiple-data) execution model of a GPU which limits their applicability to domains with simple actions that share the same code.

% \FloatBarrier
% \section{Problem Definition}
% \label{sec:problem}
% \input{04problem}

% \FloatBarrier
\section{Method}
\label{sec:methods}
\subsubsection{Problem Formulation}
Let a finite graph $\graph = (\Vertices, \Edges)$ be defined as a set of vertices \Vertices and directed edges \Edges. Each vertex $\vertex \in \Vertices$ represents a state \state in the state space of the domain \States. An edge $\ed \in \Edges$ connecting two vertices $\vertex_1$ and $\vertex_2$ in the graph represents an action $\ac \in \Aset$ that takes the agent from corresponding state $\state_1$ to $\state_2$. The action space is split into subsets of cheap (\Asetcheap) and expensive actions (\Asetexp) s.t. $\Asetcheap\cup\Asetexp = \Aset$ and corresponding cheap (\Edgescheap) and expensive edges (\Edgesexp) s.t. $\Edgescheap\cup\Edgesexp=\Edges$. An edge \ed can be represented as a pair \edge, where \state is the state at which action \ac is executed. For an edge \ed, we will refer to the corresponding state and action as $\ed.\state$ and $\ed.\ac$ respectively. \startstate is the start state and \goalreg is the goal region. $\cost:\Edges \rightarrow [0,\infty]$ is the cost associated with an edge. $\gval(\state)$ or g-value is the cost of the best path to \state from \startstate found by the algorithm so far. $\hval(\state)$ is a consistent and therefore admissible heuristic~\cite{russell2010artificial}. A path \plan is an ordered sequence of edges $\ed_{i=1}^N = \edge_{i=1}^N$, the cost of which is denoted as $\cost(\plan) = \sum_{i=1}^N \cost(\ed_i)$. The objective is to find a path \plan from $\state_0$ to a state in the goal region \goalreg with the optimal cost \costopt. There is a computational budget of \numthreads parallel threads available. There exists a pairwise heuristic function $\hval(\state, \state')$ that provides an estimate of the cost between any pair of states. It is forward-backward consistent~\cite{phillips2014pa} i.e. $\hval(\state, \state'') \leq \hval(\state, \state') + \hval(\state', \state'')~\forall~\state,\state',\state''$ and $\hval(\state, \state') \leq \costopt(\state, \state')~\forall~\state, \state'$.

\subsubsection{Algorithm}
Similar to \epase, \gepase searches over edges instead of states and exploits the notion of edge independence to parallelize this search. There are two key differences. First, \gepase handles cheap and expensive-to-evaluate edges differently. Second, 
it uses a more efficient independence check. In this section, we expand on these differences.

\begin{algorithm}[ht]
\caption{\label{alg:gepase} \wgepase: Planning Loop}
\begin{footnotesize}
\begin{algorithmic}[1]
% \State $\Aset\gets\text{ action space }$, $\numthreads \gets$ number of threads, $\graph \gets \emptyset$
% \State $\startstate\gets\text{ start state }$, $\goalreg\gets\text{ goal region}$, $terminate \gets \text{False}$
\State $terminate \gets \text{False}$
\Procedure{Plan}{}
    \State $\forall\state\in\graph$,~$\state.\gval\gets\infty$,~$\numexpanded(s)=0$
    \State $\state_0.\gval\gets0$
    \State insert $(\startstate, \dac)$ in \open
    \Comment{Dummy edge from \startstate}
    \State LOCK
    \While{$\textbf{not } terminate$}
        \If{$\open=\emptyset\textbf{ and }\be=\emptyset$}
            \State $terminate = \text{True}$
            \State UNLOCK
            \State $\Return~\emptyset$
        \EndIf
        \State remove an edge \edge from \open that has the \newline\hspace*{2.9em} smallest $\fval(\edge)$ among all states in \open that \newline\hspace*{2.9em} satisfy Equations \ref{eq:ind_check_1} and \ref{eq:ind_check_3}
        % \State $\edge \gets \open.min()$ s.t. \edge satisfies Eqs. \ref{eq:ind_check_1}, \ref{eq:ind_check_3}
        \label{alg:gepase/open_pop}
        \If{such an edge does not exist}
            \State UNLOCK
            \State wait until \open or \be change
            \label{alg:gepase/wait}
            \State LOCK
            \State continue
        \EndIf
        \If{$\state \in \goalreg$}
            \State $terminate = \text{True}$
            \State UNLOCK
            \State $\Return~\textsc{Backtrack(\state)}$
            \label{alg:gepase/construct_path}
        \EndIf
        \State UNLOCK
        \While{\edge has not been assigned a thread}
            \For{$i=1:\numthreads$}
                \If{thread $i$ is available}
                    \If{thread $i$ has not been spawned}
                        \State Spawn $\textsc{EdgeExpandThread}(i)$
                        \label{alg:gepase/spawn}
                    \EndIf
                    \State Assign \edge to thread $i$
                    \label{alg:gepase/assign_edge}
                \EndIf
            \EndFor
        \EndWhile
        \State LOCK
    \EndWhile 
    \State $terminate = \text{True}$
    \State UNLOCK
\EndProcedure
\end{algorithmic}
\end{footnotesize}
\end{algorithm}

\begin{algorithm}[ht]
\caption{\label{alg:gepase_expand} \wgepase: Edge Expansion}
\begin{footnotesize}
\begin{algorithmic}[1]
\Procedure{ExpandEdgeThread}{$i$}
    \While{$\textbf{not } terminate$}
        \If{thread $i$ has been assigned an edge \edge}
            \State $\textsc{Expand}\left(\edge\right)$
        \EndIf
    \EndWhile
\EndProcedure
\Procedure{Expand}{$\edge$}
    \label{alg:gepase_expand/add_be}
    \If{$\ac = \pac$}
        \State insert \state in \be with priority $\fval(\state)$
        \For{$\ac' \in \Asetexp$}
            \State $\fval\left(\edgeprime\right) = \gval(\state) + \wh\hval(\state)$
            \State insert \edge in \open with $\fval\left(\edgeprime\right)$
            \label{alg:gepase_expand/insert_open_exp}
        \EndFor
        \State UNLOCK
        \For{$\ac' \in \Asetcheap$}
            \State $\textsc{ExpandEdge}\left(\edgeprime\right)$
            \label{alg:gepase_expand/expand_cheap}
        \EndFor
        \State LOCK
    \Else
        \State $\textsc{ExpandEdge}\left(\edge\right)$
        \label{alg:gepase_expand/expand_expensive}
    \EndIf
\EndProcedure
\Procedure{ExpandEdge}{$\edge$}
    \State $\state', \cost\left(\edge\right)  \gets \textsc{GenerateSuccessor}
    \left(\edge\right)$
    \label{alg:gepase_expand/evaluate}
    \State LOCK
    \If{$\state' \notin \closed\cup\be$~and\\ ~~~~~~~~~~~~$\gval(\state')>\gval(\state)+\cost\left(\edge\right)$}
        \State $\gval(\state') = \gval(\state) + \cost\left(\edge\right)$
        \State $\state'.parent = \state$ 
        \State $\fval\left(\pedgenext\right) = \gval(\state') + \wh\hval(\state')$
        \State insert/update \pedgenext in \open with $\fval\left(\pedgenext\right)$
    \EndIf
    \State $\numexpanded(\state)+=1$
    \If{$\numexpanded(\state) = |\Aset|$}
        \State remove \state from \be and insert in \closed
        \label{alg:gepase_expand/remove_be}
        % \State insert \state in \closed
        % \label{alg:gepase_expand/add_closed}
    \EndIf
    \State UNLOCK
\EndProcedure
\end{algorithmic}
\end{footnotesize}
\end{algorithm}

In A*, during a state expansion, all its successors are generated and are inserted/repositioned in the open list. In \epase, the open list (\open) is a priority queue of edges (not states) that the search has generated but not expanded, where the edge with the smallest key/priority is placed in the front of the queue. The priority of an edge $\ed=\edge$ in \open is $\fval\left(\edge\right) = \gval(\state) + \hval(\state)$. Expansion of an edge \edge involves evaluating the edge to generate the successor $\state'$ and adding/updating (but not evaluating) the edges originating from $\state'$ into \open with the same priority of $\gval(\state') + \hval(\state')$. Henceforth, whenever $\gval(\state')$ changes, the positions of all of the outgoing edges from $\state'$ need to be updated in \open. To avoid this, \epase replaces all the outgoing edges from $\state'$ by a single \textit{dummy} edge $(\state', \dac)$, where \pac stands for a dummy action until the dummy edge is expanded. Every time $\gval(\state')$ changes, only the dummy edge has to be repositioned. Unlike what happens when a real edge is expanded, when the dummy edge $(\state', \dac)$ is expanded, it is replaced by the outgoing real edges from $\state'$ in \open. The real edges are expanded when they are popped from \open by an edge expansion thread. This means that every edge gets delegated to a separate thread for expansion. 

In contrast to \epase, in \gepase, when the dummy edge \pedge from \state is expanded, the cheap edges from \state are expanded immediately (Line~\ref{alg:gepase_expand/expand_cheap}, Alg~\ref{alg:gepase_expand}) i.e. the successors and costs are computed and the dummy edges originating from the successors are inserted into \open. However, the expensive edges from \state are not evaluated and are instead inserted into \open (Line~\ref{alg:gepase_expand/insert_open_exp}, Alg~\ref{alg:gepase_expand}). This means that the thread that expands the dummy edge also expands the cheap edges at the same time. This eliminates the overhead of delegating a thread for each cheap edge, improving the overall efficiency of the algorithm. The expensive edges are instead expanded when they are popped from \open and are assigned to an edge evaluation thread. If $\Asetexp = \emptyset$, \gepase behaves the same as \pase i.e. state expansions are parallelized and each thread evaluates all the outgoing edges from an expanded state sequentially. If $\Asetcheap = \emptyset$, \gepase behaves the same as \epase i.e. edge evaluations are parallelized and each thread expands a single edge at a time. 

In \epase, a single thread runs the main planning loop and pulls out edges from \open, and delegates their expansion to an edge expansion thread. To maintain optimality, an edge can only be expanded if it is independent of all edges ahead of it in \open and the edges currently being expanded i.e. in set \be~\cite{mukherjee2022epase}. An edge \ed is independent of another edge $\ed'$, if the expansion of $\ed'$ cannot possibly reduce $\gval(\ed.\state)$. Formally, this independence check is expressed by Inequalities~\ref{eq:ind_check_1}~and~\ref{eq:ind_check_2}. \wepase is a bounded suboptimal variant of \epase that trades off optimality for faster planning by introducing two inflation factors. $\wh\geq1$ inflates the priority of edges in \open i.e. $\fval\left(\edge\right)=\gval(\state)~+~\wh\hval(\state)$. $\wi\geq1$ used in Inequalities \ref{eq:ind_check_1} and ~\ref{eq:ind_check_2} relaxes the independence rule. As long as $\wi\geq\wh$, the solution cost is bounded by $\wi\cdot\costopt$. These inflation factors are similarly used in \gepase to get its bounded suboptimal variant \wgepase.

\begin{align}
    \label{eq:ind_check_1}
    \begin{split}
        \gval(\ed.\state) - \gval(\ed'.\state) \leq \wi\hval(\ed'.\state, \ed.\state)\\
        \forall\ed' \in  \open~|~\fval\left(\ed'\right) < \fval\left(\ed\right)
    \end{split}
\end{align}
\vskip -0.25cm

\begin{align}
    \label{eq:ind_check_2}
    \begin{split}
        \gval(\ed.\state) - \gval(\state') \leq \wi\hval(\state', \ed.\state)~\forall\state' \in \be
    \end{split}
\end{align}
\vskip -0.5cm

\begin{align}
    \label{eq:ind_check_3}
    \begin{split}
        \gval(\ed.\state) - \gval(\state') \leq \wi\hval(\state', \ed.\state)
        ~\forall\state' \in \be ~|~\fval\left(\state'\right) < \fval\left(\ed\right)
    \end{split}
\end{align}

In \wepase, the source states of the edges under expansion are stored in a set \be. However in large graphs, \be can contain a large number of states, and performing the independence check against the entire set can get expensive. Therefore in \wgepase, \be is a priority queue of states with priority $\fval(\state) = \gval(\state) + \wh\hval(\state)$. To ensure the independence of an edge from all edges currently being expanded, it is sufficient to perform the independence check against only the states in \be that have a lower priority than the priority of the edge in \open (Inequality~\ref{eq:ind_check_3}). Independence of an edge \ed in \open from a state $\state'$ in \be s.t. $\fval(\ed.\state) \leq f(\state')$ can be shown as follows: 

\begin{align*}
    % &\fval(\ed.\state) \leq f(\state')\\
    \implies& \gval(\ed.\state) + \wh\hval(\ed.\state) \leq \gval(\state') + \wh\hval(\state')\\
    \implies& \gval(\ed.\state) - \gval(\state') \leq \wh(\hval(\state') - \hval(\ed.\state))\\
    \implies& \gval(\ed.\state) - \gval(\state') \leq \wh\hval(\state', \ed.\state)
    \leq \wi\hval(\state', \ed.\state)\\
    &\text{(forward-backward consistency and $\wh\leq\wi$)}
\end{align*}
Beyond this, the bounded sub-optimality proof of \wgepase is the same as that of \wepase~\cite{mukherjee2022epase} since the only other way in which \wgepase differs from \wepase is in its parallelization strategy.

\begin{table*}[ht]
\centering
\begin{subtable}[h]{\textwidth}
    \footnotesize
    \centering
    \resizebox{\textwidth}{!}{%
    \begin{tabular}{c|l|lll|lll|lll||lll|lll}
    \toprule
    
                                                      &   \multicolumn{10}{c||}{$\compratio = 30$} &  \multicolumn{6}{c}{$\compratio = 300$} \\\midrule
    \textbf{\domaingrid}                              &   wA*         &  \multicolumn{3}{c|}{\wpase}          &      \multicolumn{3}{c|}{\wepase}                    &    \multicolumn{3}{c||}{\wgepase}       &    \multicolumn{3}{c|}{\wepase}       &    \multicolumn{3}{c}{\wgepase}   \\\midrule
    Threads (\numthreads)                             &   1           & 5     & 10    & 50        & 5        & 10       &  50     & 5      & 10     & 50             &   5         & 10        &  50     & 5       & 10      & 50                    \\\midrule
    Mean Time ($\si{\second}$)                    &   0.81        & 0.41  & 0.27  & 0.17      & \makeblue{0.18}     & \makeblue{0.08}     &  \makeblue{0.02}   & 0.13   & 0.06   & 0.02    &   \makeblue{1.65}      & \makeblue{0.74}      &  \makeblue{0.15}   & 1.13    & 0.51    & 0.12             \\
                                                      &               & \multicolumn{3}{c|}{}     & \multicolumn{3}{c|}{}                               &  ($\downarrow 28\%$)  & ($\downarrow 25\%$)   & ($\downarrow 0\%$)   &   \multicolumn{3}{c|}{}               & ($\downarrow 32\%$)    & ($\downarrow 31\%$)   & ($\downarrow 20\%$)  \\
    Edge Evaluations                                  &   522         & 1024  & 1513  & 3797      & 485      & 500      &  582    & 531    & 560    & 738                                                                      &   484       & 494      &  534    & 518     & 529    & 700     \\
    Mean Cost                                              &   901         & 885   & 902   & 959       & 958      & 980      &  988    & 956    & 959    & 969                                                                      &   958       & 978      &  987    & 955     & 955    & 966\\
    \bottomrule
    \end{tabular}
    }
    % \caption{(\textbf{\domaingrid})  Mean planning time, mean number of edge evaluations and mean cost for \gepase and the baselines. The percentage reduction in planning time in \wgepase from the best baseline (\wepase) is indicated with $\downarrow$.}
    \label{epase_manip/tab/2d_stats}
\end{subtable}

\begin{subtable}[h]{\textwidth}
    \footnotesize
    \centering
    \resizebox{0.85\textwidth}{!}{%
    \begin{tabular}{c|lll|lll|lll|lll}
    \toprule
    \textbf{\domainmanip} ($\compratio = 20$)          &   \multicolumn{3}{c|}{wA*}       &  \multicolumn{3}{c|}{\wpase}         &      \multicolumn{3}{c|}{\wepase}                 &    \multicolumn{3}{c}{\wgepase}                    \\\midrule
    Threads                                           & 5        & 10    & 50                      & 5     & 10       & 50                   & 5     & 10        &    50            & 5         &  10                           & 50               \\\midrule
    Success (\%)                                      & 83       & 83    & 83                      & 92    & 94       & 97                   & 91    & 93        &    94            & 92        &  94                           & 97      \\
    Number of Plans                                   & 540      & 510   & 784                     & 540   & 510      & 784                  & 540   & 510       &    784           & 540       &  510                          & 784              \\
    Mean Time ($\si{\second}$)                             & 0.38   & 0.38 & 0.39         & \makeblue{0.23}  & 0.23     & 0.15   & 0.29  & \makeblue{0.21}      & \makeblue{0.15}     & 0.2($\downarrow 13\%$)   &  0.15($\downarrow 29\%$)     & 0.13($\downarrow 13\%$)            \\
    Mean Path Length                                       & 24        & 29    & 28                      & 20    & 21       & 18        & 24        & 23        &    24                        & 23                        &  23                           & 21               \\
    \bottomrule
    \end{tabular}
    }
\end{subtable}
\caption{Evaluation metrics for \gepase and the baselines for \textbf{\domaingrid} (top) and \textbf{\domainmanip} (bottom). The percentage reduction in planning time in \wgepase from the best baseline based on mean planning time (underlined) for the same thread budget is indicated with $\downarrow$. \wastar only uses a single thread but the statistics differ in \domainmanip for different threads because the set of successful problems varies based on \numthreads for the other algorithms.}
\label{epase_manip/tab/stats}
\end{table*}

% \FloatBarrier
\section{Evaluation}
\label{sec:evaluation}
\subsection{2D Grid World}
\begin{figure}[ht]
    \centering
    \includegraphics[width=\columnwidth]{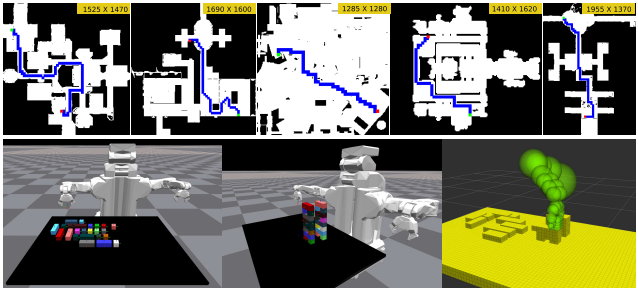}
    \caption{(Top: \textbf{\domaingrid}) 5 scaled MovingAI maps with the start state shown in green, the goal state shown in red and the computed path shown in blue. (Bottom: \textbf{\domainmanip}) The PR2 has to arrange a set of blocks on the table (left) into a given configuration (middle) with a motion planner (right) to compute \place actions.}
    \label{epase_manip/fig/nav2d_assembly}
    % \vskip -0.5cm
\end{figure}

We use 5 scaled MovingAI 2D maps~\cite{sturtevant2012benchmarks}, with state space being 2D grid coordinates (Fig.~\ref{epase_manip/fig/nav2d_assembly} top). The agent has a square footprint with a side length of 32 units. The action space comprises moving along 8 directions by 25 cell units. To check action feasibility, we collision-check the footprint at interpolated states with a 1 unit discretization. For 4 of the actions, we artificially inflate the feasibility computation to simulate the diversity in action computational effort. On average, the ratio of computation time for the actions in \Asetexp to those in \Asetcheap is $\compratio = 30$. For each map, we sample 50 random start-goal pairs and verify that there exists a solution by running \wastar. We compare \wgepase against \wastar, \wpase and \wepase. All algorithms use Euclidean distance as the heuristic with an inflation factor of 50. We see that for smaller thread budgets, \wgepase achieves $>=25\%$ lower planning times than \wepase, which is the best baseline (Table~\ref{epase_manip/tab/stats} top). However, with $\numthreads=50$, \wgepase achieves identical performance to that of \wepase. This is because in this domain, with a large number of available threads, there is no benefit of being selective about which edges should be expanded in parallel. Instead, parallelizing all edges like \wepase does is as good. However, with an additional increase in the computational cost of \Asetexp ($\compratio = 300$), \wgepase achieves a $20\%$ reduction in planning times from \wepase even for $\numthreads=50$. 

% Assembly
\subsection{\domainmanip}
We also evaluate \gepase in a manipulation planning domain for a task of assembling a set of blocks on a table into a given structure by a PR2 (Fig.~\ref{epase_manip/fig/nav2d_assembly} bottom). We collect a problem set of \place actions for 40 assembly tasks in each of which the blocks are arranged in random order on the table. \place requires a motion planner internally to compute collision-free trajectories for the 7-DoF right arm of the PR2 in a cluttered workspace to place the blocks at their desired pose. \Asetcheap comprises 22 static motion primitives that move one joint at a time by 4 or 7 degrees in either direction. \Asetexp comprises a single dynamically generated primitive that attempts to connect the expanded state to a goal configuration ($\compratio = 20$). This primitive involves solving an optimization-based IK problem to find a valid configuration space goal and then collision checking of a linearly interpolated path from the expanded state to the goal state. All primitives have a uniform cost. A backward BFS in the workspace ($x,y,z$) is used to compute the heuristic with an inflation factor of $\wh = \wi = 100$. For the problem set generation, we use \wgepase with 6 threads and a large timeout. We then evaluate all the algorithms with different thread budgets on this problem set with a timeout of $2$~\si{\second}. In the computation of the metrics (Table~\ref{epase_manip/tab/stats} bottom), we only consider the set of problems that are successfully solved and lead to a path length longer than 2 states for all algorithms. This is needed to not skew the statistics by the cases where the IK-based primitive connects the start state directly to the goal without any meaningful planning effort. We see that \wgepase consistently achieves the lowest mean planning times while maintaining a high success rate across all thread budgets. In this domain, the best baseline based on mean planning time (underlined) varies with the thread budget.

% \FloatBarrier
\section{Conclusion}
We presented \gepase, a generalized formulation of two parallel search algorithms i.e. \pase and \epase for domains with action spaces comprising a mix of cheap and expensive to evaluate actions. We showed that by employing different parallelization strategies for edges based on the computation effort required to evaluate them, \gepase achieves higher efficiency. We demonstrated this on several planning domains.

% \FloatBarrier
% \section{Acknowledgements}
% \input{09acknowledgements}

\FloatBarrier
\bibliography{main}

\end{document}